\pdfoutput=1

\documentclass[11pt]{article}

\usepackage{acl}

\usepackage{times}
\usepackage{latexsym}

\usepackage[T1]{fontenc}

\usepackage[utf8]{inputenc}
\usepackage{microtype}
\usepackage{inconsolata}
\usepackage{color, colortbl}
\definecolor{Gray}{gray}{0.9}
\usepackage{tabularx}
\usepackage{cleveref}
\usepackage{graphicx}
\usepackage{arydshln}
\small{
} 

\title{Unlocking Model Insights: A Dataset for Automated Model Card Generation}

\author{Shruti Singh \hspace{0.8em} Hitesh Lodwal \hspace{0.8em} Husain Malwat \hspace{0.8em} Rakesh Thakur \hspace{0.8em} Mayank Singh \\
Indian Institute of Technology Gandhinagar, Gujarat India \\
\texttt{\{singh\_shruti, lodwalhitesh, husainmalwat, rakesh.thakur, singh.mayank\}@iitgn.ac.in}}

\begin{document}
\maketitle
\begin{abstract}
Language models (LMs) are no longer restricted to ML community, and instruction-tuned LMs have led to a rise in autonomous AI agents. As the accessibility of LMs grows, it is imperative that an understanding of their capabilities, intended usage, and development cycle also improves. Model cards are a popular practice for documenting detailed information about an ML model. To automate model card generation, we introduce a dataset of 500 question-answer pairs for 25 ML models that cover crucial aspects of the model, such as its training configurations, datasets, biases, architecture details, and training resources. We employ annotators to extract the answers from the original paper. Further, we explore the capabilities of LMs in generating model cards by answering questions. Our initial experiments with ChatGPT-3.5, LLaMa, and Galactica showcase a significant gap in the understanding of research papers by these aforementioned LMs as well as generating factual textual responses. We posit that our dataset can be used to train models to automate the generation of model cards from paper text and reduce human effort in the model card curation process. The complete dataset is available on \href{https://osf.io/hqt7p/?view_only=3b9114e3904c4443bcd9f5c270158d37}{OSF}.
\end{abstract}

\section{Introduction}
Model cards (representative sample in \Cref{rep_model_card})  were proposed by~\citet{modelcards2019} to document the training details and intended usage of models and democratize the model development process. However, significant efforts are required to create the model card as several specific details about the model need to be extracted and organized. In recent years, the documentation of models and datasets has gained significant attention due to the rapid influx of newly proposed models and datasets. Recently, some conferences such as NeurIPS~\citep{neuripscfp} mandate the submission of datasheets for datasets, and a majority of conferences (EMNLP~\citep{emnlpcfp}, AAAI~\citep{aaaicfp}, ACL~\citep{aclcfp}) mandate discussion of reproducibility checklist, limitations and ethical considerations of the work. This highlights the importance of documenting datasets and methodology for future consumers of these artifacts. However, submission of these artifacts is restricted to specific venues and not mandated by all venues. Recently, huggingface added model cards for popular models manually~\citep{modelcardtweet}. However, manual construction of model cards is a time-consuming process, and it is difficult to manually update model cards of plethora of models submitted everyday. As a result, model cards for the majority of existing models do not exist or are incomplete. 

We propose a dataset in the format of question-answers that can be used to train models to generate model cards. We create a set of twenty general questions that seek relevant details about models (\Cref{tab:question_formulation}). We provide answers for 25 models extracted from the research papers. Our dataset can be used to train models that can extract information from research papers and automatically generate these model cards, saving time. Our dataset differs from existing QA datasets for academic papers~\citep{scienceqa2022,qasperdataset2021,pubmedqa2019,bioread2018,bioasq2015} as our question-answer pairs are specifically targeted at generating the model cards.

\begin{table}[!t]
\caption{A representative model card documenting information for the BERT-BASE-CASED model.}
\label{rep_model_card}
\small{
\begin{tabular}{|p{0.9\linewidth}|}
\hline
\rowcolor{Gray}
\textbf{Model Name: BERT-BASE-CASED} \\
\rowcolor{Gray}
Train Data: BooksCorpus (8B words) \& English Wikipedia (2.5B words) \\
\rowcolor{Gray}
Infrastructure: 4
Cloud TPUs in Pod configuration (16 TPU chips) \\
\rowcolor{Gray}
\dots \\
\rowcolor{Gray}
 Train Objective: MLM and NSP \\ \hline
\end{tabular}
}
\end{table}
\raggedbottom
\begin{figure}[!tbp]
    \centering
    \includegraphics[width=0.8\linewidth]{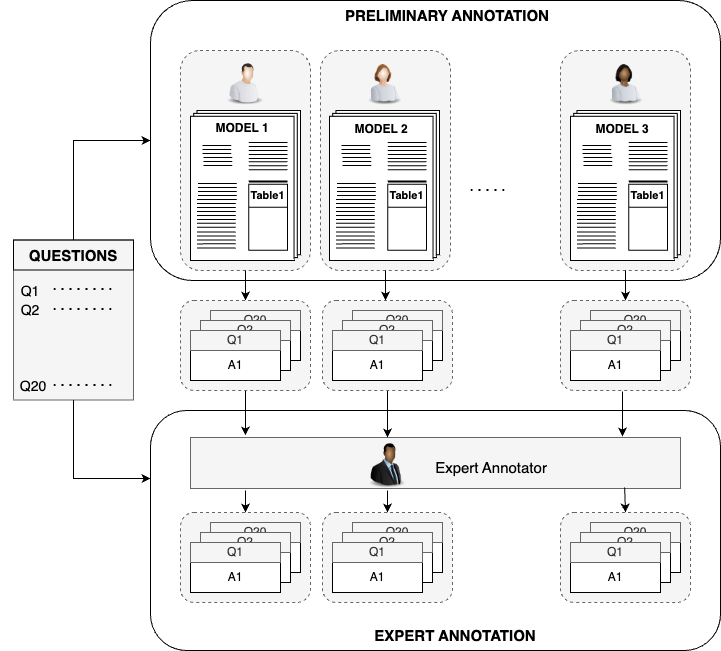}
    \caption{Twenty questions are formulated to cover model details exhaustively. The annotation pipeline consists of three stages: (i) Question Formulation, (ii) Preliminary annotation, and (iii) Expert annotation.}
    \label{fig:anno_pipeline}
\end{figure}
We evaluate large language models (LLMs) such as ChatGPT-3.5, LLaMa 7B, Galactica 125M, and Galactica 1.3B in generating the answers for model card questions in a zero-shot setting. Our experiments showcase that existing models perform poorly in generating answers. Additionally, their performance is worse if we consider factual details in the generated text. Inspected LLMs often generated memorized answers absent in the source papers and non-existent dataset names. Among the evaluated models, ChatGPT-3.5 performs the best; however, that can also be attributed to its huge size in comparison to the other probed models.
\begin{table}[!tb]
    \centering
    \small{
    \begin{tabular}{p{0.03\linewidth}p{0.9\linewidth}}
    \hline
    \rowcolor{Gray}
    \textbf{Id} & \textbf{Question} \\ \hline
    Q1 & What is the main problem statement being addressed in this paper? \\
    Q2 & What gaps in previous literature does this paper tries to address? \\ 
    Q3 & What are the main contributions of the paper? \\
    Q4 & Is the model proposed only for a specific domain, like code, images, specific text domains like finance, biomedical, etc? If yes, is it possible to extend the model to other domains? \\
    Q5 & What datasets and tasks is the model evaluated on? \\
    Q6 & Does the model show any bias/prejudice that is mentioned in paper? \\
    Q7 & List the limitations of the model discussed in the paper. \\
    Q8 & List the datasets on which the model was trained alongwith a brief summary and the size of each dataset. \\
    Q9 & List the tokenizer used alongwith the size of the vocabulary. \\
    Q10 & List the preprocessing techniques used on the dataset. \\
    Q11 & Describe the architecture details (whether it is encoder-decoder, encoder-only, or decoder-only framework, number of layers, number of heads, embedding dimension, total parameters). In case multiple models of varying sizes are trained, list details for all configurations. \\
    Q12 & Describe the training setup (e.g., learning rate, steps, epochs, optimizer, etc.) \\
    Q13 & Describe the computational resources used to train the model. \\
    Q14 & Are all details necessary to reproduce the paper provided? \\
    Q15 & What is the pretraining objective of the model?  \\
    Q16 & What is the loss function that is used to train the model? \\
    Q17 & Consider the transformer model as the base architecture for encoder-decoder (ED) models. Similarly, consider BERT and GPT as the base for encoder-only (E) and decoder-only (D) architectures. How is the architecture of this paper different from the base architectures transformer or BERT, or GPT (depending on ED, E, or D respectively)? \\
    Q18 & What experiments are conducted in the paper? Provide a brief summary of each experiment by commenting on the task description, input, expected output, and evaluation metric. \\
    Q19 & Are ablation studies conducted in the paper? If yes, which parameters are included in the ablation study? \\
    Q20 & List the future work mentioned in the paper. \\ \hline
    \end{tabular}
    }
    \caption{Question set used for extracting model card details.}
    \label{tab:question_formulation}
\end{table}

\section{Dataset}
\label{sec:dataset}
In this section, we discuss the dataset statistics, curation pipeline, and annotation strategy.
To construct our dataset, we focused on gathering ML model cards in a question-answering format. Each instance within the dataset comprises a pair consisting of a question and its corresponding answer, specifically addressing various aspects of the ML model. The questions (\Cref{tab:question_formulation}) encompass topics such as model training, architecture, problem statement, datasets used, etc. The domain of our dataset is limited to computational linguistics, and we select language models (LMs) which are successors of the transformer model~\citep{vaswani2017attention}. The dataset consists of 500 question-answer pairs for 25 models, with answers extracted from the research papers by the annotators.

The curation pipeline (\Cref{fig:anno_pipeline}) has three phases: (i) Question Formulation, (ii) Preliminary Annotation, and (iii) Expert Annotation.
The \textbf{Question Formulation} stage designs a standardized set of twenty questions that cover important aspects of the model, such as training details, architecture, problem statement, and model bias. These questions offer valuable insights into the model. We utilized the same question set for all the models included in our final collection of model card QA pairs. The complete set of questions is provided in~\Cref{tab:question_formulation}. In \textbf{Preliminary Annotation}, we curate a list of 30 popular LMs such as Longformer~\citep{beltagy2020longformer}, Transformer-XL~\citep{transformerxl2019}, BART~\citep{bart2020}, etc.; and employ 25 annotators to select a model of their preference from the model list. The 25 annotators extract the preliminary answers for different models from the respective research papers. The annotators are undergraduate and graduate students who have in-depth knowledge of traditional ML and basic knowledge of DL, including transformer architectures such as BERT. In \textbf{Expert Evaluation}, a subject expert with expertise in the field of DL (masters student in CS with prior experience in DL architectures and frameworks) reviews the answers extracted in the preliminary annotation stage. They examine the papers and assess the answers for accuracy, completeness, and relevance. Their expertise allows them to identify any inaccuracies or inconsistencies in the preliminary annotation stage answers and provide accurate assessments.
By incorporating both the preliminary and expert annotation stages, the curation pipeline adds an extra layer of annotation and combines different perspectives and expertise levels to establish a comprehensive and reliable ground truth dataset for model cards. 
All the annotators are provided with an annotated example for BERT (included \href{https://osf.io/hqt7p/?view_only=3b9114e3904c4443bcd9f5c270158d37}{here}). The annotators were instructed to extract complete answers from the research paper. The answer can span multiple sentences and paragraphs.

\section{Benchmarking LLMs for Model Card Generation}
\label{sec:benchmarkingllms}
Recent developments in instruction-following LMs~\citep{ouyang2022training,chung2022scaling,kopf2023openassistant,alpaca,vicuna2023} show improved performance for downstream tasks and has led to their increased usage in QA tasks. We evaluate the performance of LLMs in zero-shot QA for generating the model cards. ChatGPT-3.5~\citep{wu2023brief} by OpenAI uses RLHF, and its training data details are publicly unavailable. LLaMa~\citep{touvron2023llama} by Meta AI is trained on trillions of tokens from public data, including arXiv \LaTeX files, Github, and Wikipedia. Galactica~\citep{taylor2022galactica} is a scientific LM by Meta AI, trained on a 106 billion tokens data including research papers, scientific KBs, and Github (detailed discussion in appendix).
\subsection{Prompting LLMs for QA}
\label{ssec:eval_procedure}
\begin{figure}[!t]
    \centering
    \includegraphics[width=0.8\linewidth]{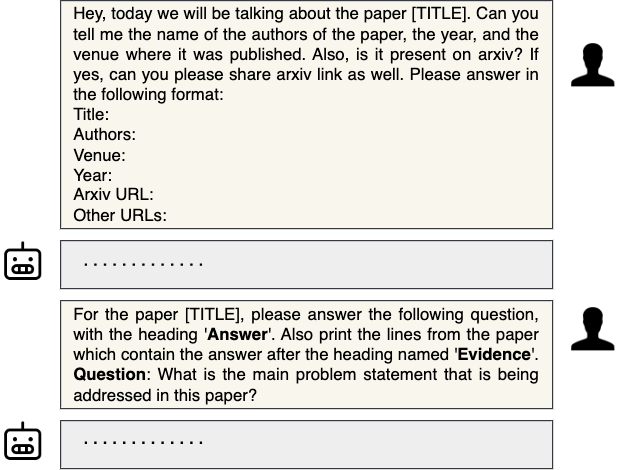}
    \caption{Prompting ChatGPT-3.5 for generating answers for model card questions.}
    \label{fig:propmtingllm}
\end{figure}

We evaluate ChatGPT-3.5, LLaMa (7B parameters), Galactica (125M parameters), and Galactica (1.3B parameters) in a zero-shot setting to test their ability to generate answers for model card questions. The subject expert annotator prompts ChatGPT for a particular model in a single session, starting with a general prompt trying to elicit model details. \Cref{fig:propmtingllm}, showcases the prompting procedure employed to generate answers. Based on the ground truth, the subject expert also marks the ChatGPT-3.5 answer into correct and incorrect spans. For LLaMa evaluation, we download the original LLaMa 7B weights made available by Meta AI and then use the llama.cpp repository~\citep{llamacpp} to convert it to a 4-bit integer quantized model. We directly use the Galactica 125M and Galactica 1.3B models made available by Meta AI to generate the answers. For LLaMa and Galactica models, we restrict the output size to 1000 tokens. 

\subsection{Qualitative and Quantitative Evaluation}
\label{ssec:eval_metrics}
We perform a quantitative evaluation of the LLM-generated responses by computing the BLEU~\citep{papineni2002bleu}, ROUGE-L~\citep{lin-2004-rouge}, and  BERT-Score~\citep{bert-score}.  We represent the BERT-Score Precision, Recall, and F1 Score as BS-P, BS-R, and BS-F1, respectively. However, these automatic evaluation metrics do not take into account the factuality of the answers. For e.g., an LLM-generated answer which exactly matches the ground truth answer but differs in training dataset name receives a high score, disregarding the inaccurate dataset name, a crucial aspect in an answer describing the training dataset. For e.g., the mBART~\citep{mBART2020} model is evaluated on the WMT-16, WMT-17, WMT-18, and WMT-19 datasets, but LLaMa generated answer says \texttt{`mBART achieves state-of-art accuracy on WMT-14 dataset'}, which is incorrect and not mentioned in the paper.

To better understand the factuality of generated answers, we do a small-scale qualitative evaluation. An expert annotator (different from the annotator in data curation) evaluates the LLM-generated answers for five models in the dataset for each of the LLMs (ChatGPT-3.5, LLaMa, Galactica 125M, and Galactica 1.3B). We choose three different labels, Completely Correct (CC), Partially Correct (PC), and Incorrect (IC); to evaluate the answers. If the LLM answer exactly matches the ground truth answer in terms of relevance, factuality, and exhaustively covers all crucial facts, the response generated by the LLM is marked CC. If the generated response consists of some correct and some incorrect facts, it is marked PC. If none of the facts in the generated answer are correct, the answer is marked IC. Special attention is paid to dataset, model, and task names during evaluation, and any mismatch in the names is marked as incorrect. In the earlier provided example, where LLaMa listed WMT-14 dataset while the ground truth was WMT-16, WMT-17, WMT-18, and WMT-19 datasets, the subject expert annotator marked the answer as incorrect, even though the phrase `WMT' will be matched by other automatic evaluation metrics. BS-F1 for these pair of sentences is high, 0.795, even though the dataset is incorrect. We present representative examples of CC, PC, and IC in~\Cref{tab:qual_eval_examples}.
\begin{table}[!tp]
\centering
\small{
\begin{tabular}{p{0.4\linewidth}p{0.4\linewidth}p{0.05\linewidth}}
\hline
\rowcolor{Gray}
\textbf{LLM-generated} & \textbf{Ground Truth} & \textbf{Label}   \\  \hline
The architecture is based on the \underline{BERT} architecture, an encoder-only model. & DistilBERT has the same general architecture as \underline{BERT}. &   CC    \\ \hdashline
The pretraining objective is same as BERT, i.e., \underline{predict some masked tokens}.  &  We follow the BERT training, i.e. \underline{MLM} but without the \underline{next sentence prediction objective}. &   PC    \\ \hdashline
The model achieves state-of-art accuracy on \underline{WMT-14 dataset}. &  Results indicate a new SOTA on the \underline{WMT-16 English-Romanian}. &   IC    \\  \hline
\end{tabular}
}
\caption{Representative examples of LLM-generate text and ground truth, marked as CC, PC, and IC. Artifact names and concepts (underlined) are crucial in determining the label.}
\label{tab:qual_eval_examples}
\end{table}
\begin{table*}[!tbhp]
\centering
\setlength{\tabcolsep}{2pt}
\small{
\begin{tabular}{lllllllllllllllllllll}
\hline
\rowcolor{Gray}
          & Q1 & Q2 & Q3 & Q4 & Q5 & Q6 & Q7 & Q8 & Q9 & Q10 & Q11 & Q12 & Q13 & Q14 & Q15 & Q16 & Q17 & Q18 & Q19 & Q20  \\ \hline
ChatGPT-3.5 &  \underline{0.15} & 0.14 & \underline{0.18} & \underline{0.23} & \underline{0.15} & \underline{0.3} &\underline{ 0.18 }& 0.24 & \underline{0.19} & \underline{0.19} & \underline{0.23} & \underline{0.22} & \underline{0.27} & \underline{0.22} & \underline{0.21} & \underline{0.2} & 0.23 & \underline{0.19} & \underline{0.17} & \underline{0.21} \\
LLaMa 7B  &  \underline{0.15} & \underline{0.23} & \underline{0.18} & 0.2 & 0.1 & 0.15 & 0.14 & 0.21 & 0.17 & 0.2 & 0.19 & 0.17 & 0.17 & 0.06 & 0.15 & 0.1 & \underline{0.26} & 0.17 & 0.13 & 0.19 \\
Galactica 125M &  0.11 & 0.14 & 0.1 & 0.11 & 0.09 & 0.07 & 0.11 & 0.25 & 0.1 & 0.12 & 0.17 & 0.1 & 0.09 & 0.04 & 0.11 & 0.05 & 0.18 & 0.08 & 0.05 & 0.11 \\
Galactica 1.3B &  0.2 & 0.18 & 0.16 & 0.16 & 0.13 & 0.13 & 0.14 & \underline{0.27} & 0.15 & 0.14 & 0.22 & 0.16 & 0.18 & 0.09 & 0.1 & 0.09 & 0.22 & 0.13 & 0.13 & 0.14 \\ \hline
\end{tabular}
}
\caption{ROUGE-L scores of the LLMs in generating the answers for the model card for each question.}
\label{tab:eval_scores_quewise}
\end{table*}

\begin{table}[!tbp]
\centering
\setlength{\tabcolsep}{3pt}
\small{
\begin{tabular}{llllll}
\hline
\rowcolor{Gray}
               & BLEU & RO-L & BS-P & BS-R & BS-F1 \\ \hline
ChatGPT-3.5  & \underline{0.0230}  &  \underline{0.2052} &  \underline{0.5785}           &  \underline{0.5638}  & \underline{0.5679} \\
LLaMa 7B & \textbf{0.0083}  &  \textbf{0.1958}  &  \textbf{0.4972}   &   \textbf{0.5298}  &  \textbf{0.5086} \\
Galactica 125M &  {0.0079} &  0.1417 &   0.4213  &  0.5004   &  0.4493  \\
Galactica 1.3B & 0.0072 &  0.1559 & 0.3981  &   0.5112  &  0.4374  \\ \hline
\end{tabular}
}
\caption{Scores of the LLM-generated answers for the model card questions using BLEU, ROUGE-L (RO-L), BS-P, BS-R, and BS-F1. Best scores are \underline{underlined}, and second-best are in \textbf{bold}.}
\label{tab:eval_scores}
\end{table}
\begin{table}[!tbp]
\centering
\small{
\begin{tabular}{llll}
\hline
\rowcolor{Gray}
               & \textbf{CC} & \textbf{PC} & \textbf{IC} \\ \hline
ChatGPT-3.5        & 25.0 & 43.0  & 32.0   \\
LLaMa          &  0  & 17.0  &  83.0 \\
Galactica 125M & 0   & 8.0 & 92.0   \\
Galactica 1.3B &  0  & 15.0  & 85.0   \\ \hline
\end{tabular}
}
\caption{Percentage CC, PC, and IC answer evaluated on five ML models (20 questions each, total 100 QA pairs). The majority of LLM-generated answers are IC or PC.}
\label{tab:qual_scores}
\end{table}
\raggedbottom
\subsection{Results of zero-shot QA using LLMs}
\label{ssec:eval_results}
We showcase the average BLEU, ROUGE-L, and BERT Score P, R, and F1 in~\Cref{tab:eval_scores}.
We observe that ChatGPT-3.5 performs the best across all metrics, and LLaMa 7B performs comparably to ChatGPT-3.5. It should be noted that the models have different parameters, and hence the results should not be generalized to bigger LLaMa and Galactica models. Question-wise scores (\Cref{tab:eval_scores_quewise}) show that ChatGPT-3.5 achieves the highest ROUGE-L scores in answering Q6 and Q13, questions about model biases and computational resources. Galactica 125M performed worst than all models, often generating repetitive text, which could also be attributed to its size. A major limitation of all LLMs is that their text generation is not grounded on facts. To analyze the LLM-generated answers for factuality, we present the results for qualitative evaluation in~\Cref{tab:qual_scores}. The low scores indicate that most of the generated answers are incorrect and often memorized based on the frequency in the training dataset. For example, a manual analysis reveals that ChatGPT-3.5 often reports the training infrastructure as 3 V100 GPUs, irrespective of the model for which we probe for the training details. While ChatGPT-3.5 generated mostly partially correct responses, other models' responses were often incorrect, with Galactica 125M having almost 92\% incorrect responses.

\section{Potential Usage}
\label{sec:usage}
We posit that our dataset of QA pairs can be used to train models for generating model cards. Our dataset can be used for instruction tuning LLMs for generating ML model cards by prompting specific questions about different aspects of the model. Similarly, our dataset of answers generated from LLaMa and Galactica models can be utilized as 
there are initial answers from LLMs and ground truths. So LLMs can be prompted with the whole conversation, where they are provided with a question, incorrect answer, and then pointed the correct answer. For ChatGPT-3.5-generated responses, we provide another layer where we label which spans of the answer are incorrect.

\section{Related Work}
\label{sec:rel_work}
\textbf{Information Seeking Datasets for Research Papers:} Previous works curate datasets from research papers for fact verification and facilitating paper reading. 
ScienceQA~\citep{scienceqa2022} comprises 100k synthetic question-answer-context triples, where the questions are generated from a filtered set of noun phrases extracted from 1825 IJCAI papers. QASPER~\citep{qasperdataset2021} is a dataset of 5049 questions over 1585 NLP papers, with  extractive, abstractive, and binary yes/no answers. 
Multiple works curate QA datasets in the biomedical domain, such as BIOASQ~\citet{bioasq2015}, BioRead~\citet{bioread2018}, and PubMedQA dataset~\citep{pubmedqa2019}.
\textsc{SciFact}~\citep{waddenscifact2020} dataset consists of 1.4k expert-written biomedical scientific claims and evidence abstracts with labels (support or refutes) and rationales.  
In comparison to the existing datasets, our dataset is highly specific and curates model card information for an efficient understanding of ML models.

\textbf{Model cards and allied concepts:} Similar to Model cards~\citep{modelcards2019}, datasheets for datasets~\citep{gebru2021datasheets} are proposed to document dataset information and bridge the gap between dataset creators and consumers. Datasheets are proposed to document the data collection process, sources, intended use cases, etc., to promote transparency. \textsc{RiskCards}~\citep{riskcards2023} document the risks associated with LMs by constructing a detailed skeleton to document risks such as harm type (which group is at what type of harm), references, and sample prompts and LM output. AI Usage Cards~\citep{aiusagecards2023} is proposed to standardize the reporting of the usage of AI technologies such as ChatGPT-3.5. These works emphasize the need for documenting artifacts in this age of information overload. We posit that our dataset will be utilized by models to assist in automating model card generation. This has positive prospects as instruction tuning has shown promise in improving the ability of LLMs to follow instructions and show significant improvement in downstream tasks. Apart from our QA dataset, we also provide the QA pairs generated by various LLMs such as ChatGPT-3.5, LLaMa, and Galactica. Lastly, model cards for some models are available on huggingface; however, those are unstructured and available in free-form text. Our dataset provides structured model card details in a QA format that can be leveraged by LLMs to learn model card generation. There are no ethical concerns with our dataset and it doesn't contain offensive content or personally identifiable information.

\section{Conclusion}
\label{sec:conclusion}
To summarize, we curate a QA dataset of 500 pairs from research papers of 25 machine learning models. A set of 20 general questions are curated for the purpose of seeking crucial details of model training, dataset, problem statement, ablation studies, etc. We employed a two-stage annotation pipeline consisting of preliminary and expert annotation stages to ensure the quality of the curated dataset. Next, we evaluate the capability of LLMs ChatGPT-3.5, LLaMa, and Galactica in generating answers for the model card questions. All the evaluated models include infactual details in the answer, highlighting the potential for developing better models for model card generation. 
In the future, we plan on expanding the question set and covering models from other domains, such as CV and robotics.

\bibliography{anthology,custom}

\appendix
\section{Appendix}
\label{sec:appendix}
\subsection{Dataset Statistics}
Our dataset consists of 500 examples, i.e. twenty question-answer pairs for 25 language models. The list of language models is presented in~\Cref{tab:llmslist}.
\begin{table}[]
    \centering
    \begin{tabular}{c|c|c}
        \hline
        \rowcolor{Gray}
        \multicolumn{3}{c}{Language Models} \\ \hline
        MBART & DistilBERT & Longformer \\
        Sparse Transformer & ERNIE & PEGASUS \\
        Reformer & BART & RoBERTa  \\
        BigBird & MobileBERT & GPT2  \\
        ELECTRA & StructBERT & MuRIL  \\
        GPT & SpanBERT & ALBERT \\
        XLNet & BERT-PLI & Transformer \\
        T5 & FNet & TransformerXL \\
        SciBERT \\ \hline
    \end{tabular}
    \caption{List of language models covered in the model card dataset.}
    \label{tab:llmslist}
\end{table}

\begin{table}[!htbp]
    \centering
    \caption{The dataset consists of 20 QA pairs for 25 models, amounting to overall 500 QA pairs for model card generation.}
    \begin{tabular}{ll}
    \hline
    Models & 25 \\
    Questions per model & 20 \\ 
    Total QA pairs & 500 \\ \hline
    \end{tabular}
    \label{tab:data_stats}
\end{table}

\subsection{Benchmarking LLMs for Model Card Generation}
\begin{figure}[!htbp]
    \centering
    \small{
        \includegraphics[width=0.9\linewidth]{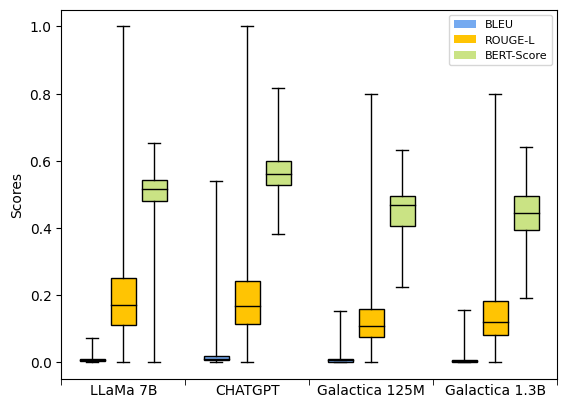}
        \caption{Distribution of BLEU, ROUGE-L, and BS F1 for ChatGPT-3.5, LLaMa 7B, Galactica 125M, and Galactica 1.3B.}
        \label{fig:metric_box_plots}
    }
\end{figure}

In~\Cref{fig:metric_box_plots}, we present the box plots of scores of the LLaMa 7B, ChatGPT, Galactica 125M, and the Galactica 1.3B model. We compare the BLEU, Rouge-L, and BERT-Score of the evaluated models. The plot indicates that the median ROUGE-L scores for LLaMa 7B and the ChatGPT model are better than the Galactica models. ChatGPT outperforms the other models in terms of BERT-Score. The BLEU scores for all the models denote poor performance, with scores almost close to zero, indicating significant scope for improving these models for the model card generation task. 

\subsection{Large Language Models}
\textbf{ChatGPT-3.5}~\citep{wu2023brief}, developed by OpenAI, is a sibling to the InstructGPT~\citep{ouyang2022training}, which uses human feedback to align outputs of language models with the human intent. While the training dataset detail of ChatGPT-3.5 is not publicly available, the model is trained with reinforcement learning with human feedback on a dialogue dataset. \\
\textbf{LLaMa} (Large Language Model Meta AI)~\citep{touvron2023llama} is an LLM developed by Meta AI in October 2021, available in sizes ranging from 7B to 65B parameters. LLaMa is trained on trillions of tokens, using publicly available datasets such as C4, CommonCrawl, Github, Wikipedia, Gutenberg and Books3, arXiv \LaTeX files, and StackExchange. \\
\textbf{Galactica}~\citep{taylor2022galactica} is a scientific language model by Meta AI, trained on a 106 billion tokens dataset. Galactica is trained on a diverse dataset of research papers, reference materials from encyclopedias, textbooks, Wikipedia, StackExchange, scientific knowledge bases, scientific and academic Common Crawl, and academic Github code repositories.

\end{document}